\documentclass[journal,twoside,web]{ieeecolor}
\usepackage{jsen}
\usepackage{cite}
\usepackage{amsmath,amssymb,amsfonts}
\usepackage{algorithmic}
\usepackage{graphicx}
\usepackage{textcomp}
\usepackage{wrapfig}
\usepackage{bm}

\def\BibTeX{{\rm B\kern-.05em{\sc i\kern-.025em b}\kern-.08em
    T\kern-.1667em\lower.7ex\hbox{E}\kern-.125emX}}
\definecolor{abstractbg}{rgb}{0.89804,0.94510,0.83137}
\begin{document}
\title{Noise Analysis and Hierarchical Adaptive Body State Estimator For Biped Robot Walking With ESVC Foot}
\author{Boyang Chen, Xizhe Zang, Chao Song, Yue Zhang, Xuehe Zhang and Jie Zhao
\thanks{This work was supported in part by the Major Research Plan of the National Natural Science
Foundation of China (92048301).}
\thanks{Boyang Chen, Xizhe Zang, Chao Song, Yue Zhang, Xuehe Zhang and Jie Zhao are with the School of Mechatronics Engineering, Harbin Institute of Technology, Harbing 150001, China(e-mail:chenboyang@stu.hit.edu.cn,zangxizhe@hit.edu.cn, songchao@stu.hit.edu.cn,yzhanghit@stu.hit.edu.cn, zhangxuehe@hit.edu.cn,jzhao@hit.edu.cn).}
\thanks{Corresponding authors: Xuehe Zhang.}
}

\IEEEtitleabstractindextext{%
\fcolorbox{abstractbg}{abstractbg}{%
\begin{minipage}{\textwidth}%
\label{abstractfig}
\begin{wrapfigure}[16]{h}{3.2in}%
\includegraphics[width=3.2in]{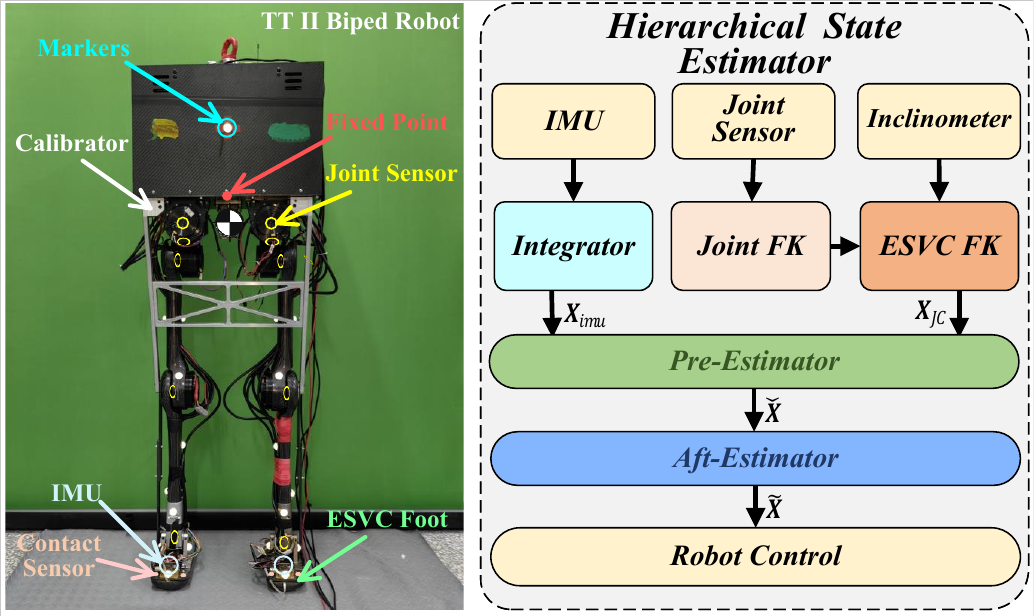}%
\end{wrapfigure}%
\begin{abstract}
The ESVC(Ellipse-based Segmental Varying Curvature) foot, a robot foot design inspired by the rollover shape of the human foot, significantly enhances the energy efficiency of the robot walking gait. However, due to the tilt of the supporting leg, the error of  the contact model are amplified, making robot state estimation more challenging. Therefore, this paper focuses on the noise analysis and state estimation for robot walking with the ESVC foot. First, through physical robot experiments, we investigate the effect of the ESVC foot on robot measurement noise and process noise. and a noise-time regression model using sliding window strategy is developed. Then, a hierarchical adaptive state estimator for biped robots with the ESVC foot is proposed. The state estimator consists of two stages: pre-estimation and post-estimation. In the pre-estimation stage, a data fusion-based estimation is employed to process the sensory data. During post-estimation, the acceleration of center of mass is first estimated, and then the noise covariance matrices are adjusted based on the regression model. Following that, an EKF(Extended Kalman Filter) based approach is applied to estimate the centroid state during robot walking. Physical experiments demonstrate that the proposed adaptive state estimator for biped robot walking with the ESVC foot not only provides higher precision than both EKF and Adaptive EKF, but also converges faster under varying noise conditions.
\end{abstract}

\begin{IEEEkeywords}
Adaptive state estimation, biped locomotion, curvature foot, noise measurement.
\end{IEEEkeywords}
\end{minipage}}}

\maketitle

\section{Introduction}
\label{sec1}
\IEEEPARstart{B}{ipedal} robot technology has made significant progress in recent years, such as G1 from Unitree\cite{b1}, Atlas of Boston Dynamics\cite{b2} and Digit designed by Agility Robotics\cite{b3}. In laboratory environment, robots have already demonstrated impressive motion capabilities that are close to, or even surpass, those of humans. However, from the prospective of practical application, numerous challenges still exist, among which gait energy efficiency and ground adaptability are particularly important. When excluding upper limb operations, the foot, as the only part interacting with the external environment, directly affect the evolution of the entire robot system. Therefore, to improve bipedal walking performance, various foot designs for different scenarios have been investigated and implemented\cite{b4}\cite{b5}\cite{b6}\cite{b7}.

Among these studies, a key question is that how to effectively integrate foot design with robot locomotion and fully exploit the performance enhancements provided by the meticulous foot designs. This question is challenging and a prerequisite is that how the foot designs influence locomotion dynamics. In particular, the impact of the foot designs on walking-state variations needs careful consideration. In other words, from a feedback control perspective, understanding the effect of foot design on system noise and ensuring accurate state estimation are indispensable.

In terms of rollover shape, mainstream robot foot designs can be categorized as: arched foot\cite{b8}, point foot\cite{b9}, flat foot\cite{b10}, line foot\cite{b11}, arc/curved foot\cite{b6}\cite{b12}, bionic foot\cite{b13} and flexible heterogeneous foot\cite{b14}. Among them, point foot, line foot, and flat foot are the most widely used. The point foot, with negligible mass, maintains point contact with the ground at all times in both the sagittal and coronal planes, greatly simplifying the modeling and control of biped robots, but it cannot provide sufficient frictional torque. The line foot maintains line contact in the sagittal plane and point contact in the coronal plane, compensating for the lack of friction of point foot. However, its energy efficiency is much lower than that of arc/curved or bionic designs. Moreover, line foot and point foot lack sufficient sensory equipment so the foot data and body state have to be estimated through additional methods. The flat foot can compensate for the shortcomings of both; however, maintaining full contact requires the supporting foot to remain parallel to the ground at all times. This necessitates actuation of the ankle joint in both the pitch and roll directions, which in turn hinders the mass centralization of the robot. Additionally, the impact energy dissipation caused by the contact also results in lower gait efficiency. The bionic foot and flexible heterogeneous foot are not discussed in this paper due to their complexity and specific application scenarios.

In fact, it is difficult to find a generalized principle for robot foot design currently, especially when balancing multiple design criteria. Therefore the rollover shape of human foot serves as a good bionic template. Researchers in prosthetics and human anatomy have found that the bionic rollover shape can be divided into three parts: the fore-foot, mid-foot, and hind-foot. Each segment exhibits a varying curvature, with the heel and toes being more curved, while the mid segment remains relatively flat. The segmented varying curvature configuration allows for higher energy efficiency during human biped walking\cite{b15} \cite{b16}\cite{b17}. Inspired by this an ellipse based segmental varying curvature foot (ESVC foot) has been proposed in our previous work\cite{b12}. Multiple elliptical arcs are engaged to approximate the configuration of the rollover shape of human foot. Physical experiments on TT II robot illustrate that the foot design significantly improves the gait energy efficiency, especially when there exists lateral velocity. During walking, the support leg of the robot with ESVC foot undergoes a passive rotation in coronal plane and the contact point will rollover along the elliptical arcs. As the analytical form of the contact model does not exist due to the incomplete elliptic integral, an approximation contact model based on elementary functions is provide in our previous research. However, despite the simplicity of this method, the modeling error cannot be eliminated, and the error is amplified by the support leg rotation during robot walking. This means that directly using joint state data to obtain the robot body state is too coarse and sensitive for feedback control. Hence, this paper we try to reveal how the ESVC foot affect noise distribution of the robot system and focuses on the state estimation with ESVC foot.

The state estimation of bipedal robot has long been a focus of researchers, as it serves as a crucial foundation for feedback control. Due to the high dimensional non-linearity of the legged robot system, the extended Kalman filter(EKF) based method is widely used for state estimation\cite{b18}, \cite{b19}, \cite{b20}, \cite{b21}, \cite{b22}. By setting the process noise and measurement noise covariance as constant, Chen et al.\cite{b19} used the EKF to estimate the state of a fixed point(FP). The center of mass(CoM) state is then approximated through the FP and incorporated into feedback control based on the linear inverted pendulum model. This method is suitable for slower robot walking, where the FP behaves similarly to the CoM. Kang et.al.\cite{b20} proposed a nonlinear distributed state estimation framework for legged robot system. Through EKF and moving horizon estimation, the linear and angular states of the robot floating base can be estimated decoupledly, while maintaining a balance between accuracy and computational efficiency is feasible. Gao et al.\cite{b22} proposed a robust adaptive state estimator based on the Invariant Extended Kalman Filter(IEKF), and by segmenting the design of adaptive adjustment factors, the estimator can adjust the noise covariance dynamically, and in their physical experiments, the velocity estimation of CoM demonstrated around 50\% improvement compared to the original IEKF. Kim et.al.\cite{b23} redefined the state estimation as a Maximum A Posteriori (MAP) problem, achieving estimation results comparable to the IEKF under different contact conditions. Youm et.al.\cite{b24} proposed a learning-based approach that combines the Neural Measurement Network with EKF, significantly reducing position drift. The network is trained using only simulation data.

This paper focuses on the impact of the ESVC foot to the system noise and state estimation of walking bipedal robots The contributions are as follows: first, we investigate the impact of the ESVC foot and contact model error on the robot measurement noise and process noise through physical experiments. Then a temporal regression model for noise covariance is developed using sliding window strategy and weighted cubic regression. Finally, we propose a hierarchical adaptive state estimator based on the temporal model. Under feedback control using the Hybrid Linear Inverted Pendulum (HLIP)\cite{b25}, the proposed state estimator not only accurately provides body state data during robot walking, but also converges rapidly when the walking gait varying, compared to the classical EKF and the adaptive variants with fixed noise covariance.

The organization of this paper is as follows: In Section II.a, we review the ESVC foot model and demonstrate how to obtain the sensory data of robot state through the approximation model. In Section II.b, we first reveal the distribution of measurement noise and process noise through fixed point trajectory tracking experiments and physical walking experiments, respectively. Then the temporal  regression model is designed for the initial guess of  estimation. In Section III, we discuss the hierarchical adaptive state estimator. Section IV presents the validation and evaluation of the proposed state estimator. Conclusions and future work are presented in Section V. 

\section{Noise Analysis And Approximation}
\label{sec2}
\subsection{Sensory State Data Acquisition}
\label{sec2.1}
The quality of sensory data directly impacts the body state estimation, especially when errors are introduced by the varying contact point model of the ESVC foot. In order to investigate the impact, we equipped the ESVC foot to bipedal robot TT II as shown in the abstract figure\ref{abstractfig}. To acquire the sensory body state data relative to the contact frame, including the center of mass and the swing foot, six virtual joints are typically introduced to describe the robot motion with respect to the inertial frame which allows the body state can be calculated in a manipulator-like manner. However, this method is susceptible to sensory noise and also increases the computational load on the onboard controller. In our previous work\cite{b12}, we derived the spatial transformation matrices of foot upper center frame, foot fixed frame and contact point frame. Through the ESVC kinematics, we divide the configuration space into the internal space and the external space and the robot body state can be computed effciently and decoupledly within the two spaces. In the internal space, the robot can be regarded as being fixed to the ground at the upper center frame of the support foot, making the sensory body state data only dependent on the joint states:
\begin{equation}
	\label{eq1}
	\textbf{X}^{O_i} = f\left( \textbf{q}_i, \dot{\textbf{q}}_i\right).
\end{equation}
For simplicity, we use $O_i$ and $C$ to denote the upper center frame of the support foot and contact frame, respectively. Then, in the external space, the sensory body state from joint space data calculation $\textbf{X}_{JC}={[\textbf{p}_{JC}^C,\textbf{v}_{JC}^C]}^T$ with respect to the inertial frame can be obtained through the  matrix ${}_{O_i}^C\!\bm{T}_{E}$ and its derivatives ${}_{O_i}^C\!\dot{\bm{T}}_{E}$, which represents the transformation between frame $O_i$ and frame $C$ of ESVC foot, as shown in Eq.\ref{eq2}. Since the transformation matrix is based on elementary functions and the derivatives can be obtain directly through numerical differencing, the calculation of the external space is simplified to a single matrix multiplication:
\begin{equation}
	\label{eq2}
	\left[\bm{p}_{JC}^C, 1, \bm{v}_{JC}^C, 1\right]^T=
		\begin{bmatrix}
			{}_{O_i}^C\!\bm{T}_{E}&0\\
			{}_{O_i}^C\!\dot{\bm{T}}_{E}&{}_{O_i}^C\!\bm{T}_{E}
		\end{bmatrix}
		\left[\bm{p}^{O_i}, 1, \bm{v}^{O_i}, 1\right]^T
\end{equation}

TT II robot is equipped with three groups of IMU and inclinometers, as shown in the top-right subfigure of the abstract figure. As two sensor groups are mounted on the surface of the ESVC foot to directly provide the states of both the swing and support foot, the matrix ${}_{O_i}^C\!\bm{T}_{E}$ and ${}_{O_i}^C\!\dot{\bm{T}}_{E}$ in Eq.\ref{eq2} can be calculated accordingly. The third group is mounted at a fixed position(FP) which is close to the CoM when standing, to obtain approximate state data of the CoM $\textbf{X}_{imu}=[\textbf{p}_{imu}, \textbf{v}_{imu}]^T$. In other words, the TT II robot has two separate sources of sensory body state data, $\textbf{X}_{JC}$ and $\textbf{X}_{imu}$.

\subsection{Noise Analysis And Regression Fitting}
\label{sec2.2}
In general, measurement noise and process noise are assumed to follow a Gaussian distribution, and their covariance matrices are often set as constant in state estimation. However, due to the complexity of bipedal system, the process covariance matrix often varies over time. Moreover, when the robot is equipped with the ESVC foot, its model error also affects the noise distribution. Therefore, in this section, we analyze the noise characteristics and determine the distribution of the robot with the ESVC foot through two experiments.

\subsubsection{Analysis of Measurement Noise}
\label{sec2.2.1}
Measurement noise is composed of the multiple sensors noise. For the two separate sources of sensory body state data of TT II, we use $\bm{\sigma}^2_{imu}=[\bm{\sigma}^2_{posimu},\bm{\sigma}^2_{velimu}]$ and $\bm{\sigma}^2_{JC}=[\bm{\sigma}^2_{posJC},\bm{\sigma}^2_{velJC}]$ to denote the measurement variance. The target trajectory of FP is designed to follow a path resembling the Lemniscate of Bernoulli. In the joint space, the actuators are controlled to accurately track the target joint position trajectory generated by the ESVC inverse kinematics. Several marker points are placed to indicate the positions of various body links of the robot, and a depth camera is employed to capture the joint angles.

To minimize the influence of mechanical clearances and motor backlash on measurement noise, the FP target trajectory is constrained to the coronal plane. As a result, the feet exhibit displacement only along the Y-axis and rotation about the X-axis. Although the foot roll angle is relatively small, its effect on the FP point cannot be neglected due to amplification caused by the supporting leg length. Finally, the position and velocity of the FP relative to the fixed frame are computed as ground truth using the URDF model, with two additional virtual joints introduced of the motion of Y-axis and X-axis.
\begin{figure}[!t]
	\centerline{\includegraphics[width=\columnwidth]{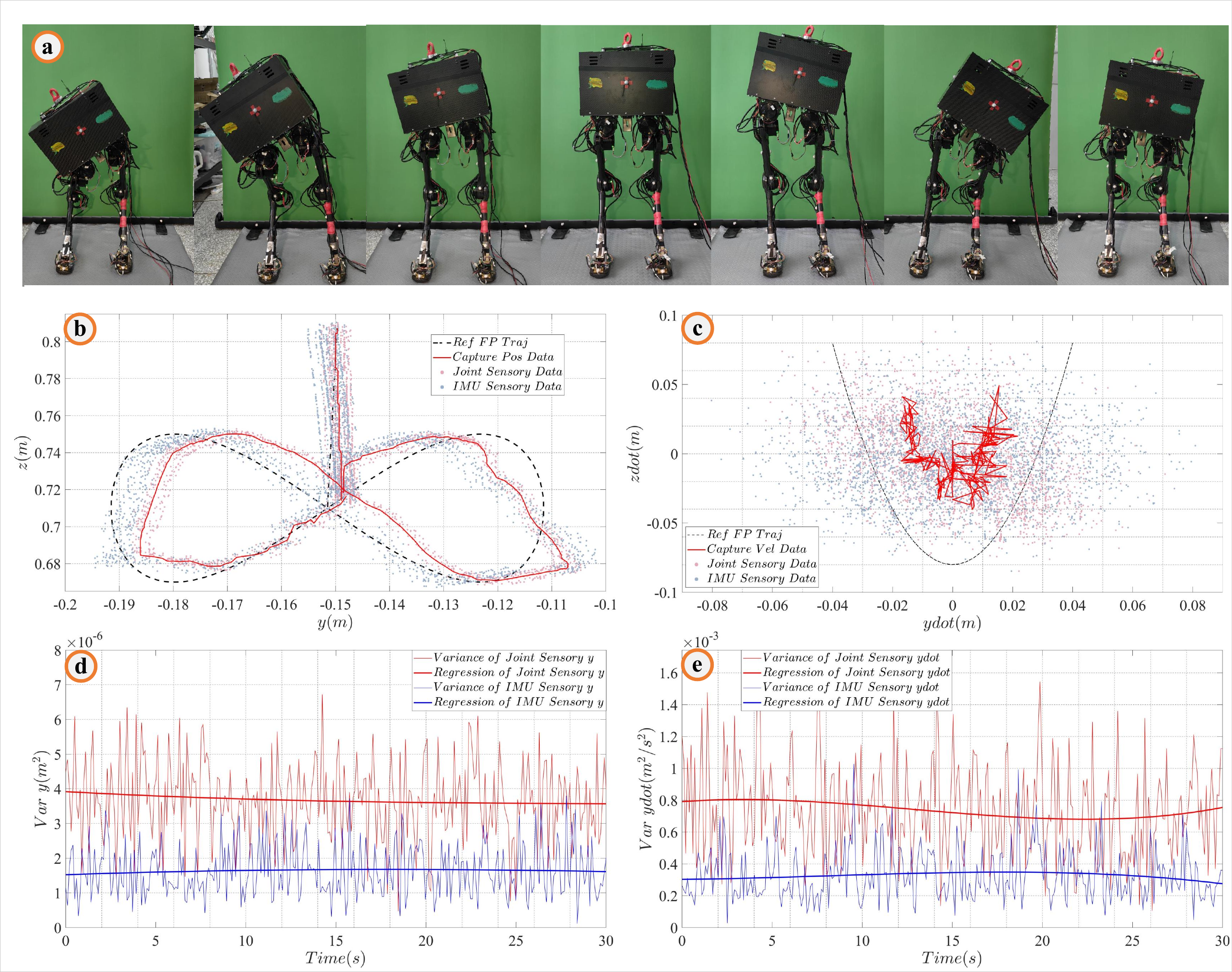}}
	\caption{(a). Snapshots FP trajectory tracking physical experiments on robot TT II. (b). Coronal FP position trajectory over ten experiments. (c). Coronal FP velocity trajectory over ten experiments. (d). FP position error variance distribution. (e). FP velocity error variance distribution.}
	\label{fig_a}
\end{figure}

Snapshots of the robot physical experiments are shown in Fig\ref{fig_a}.a. To eliminate random errors, the same experiment was repeated ten times, and the position and velocity of the FP in the coronal plane are depicted in Fig\ref{fig_a}.b and Fig\ref{fig_a}.c. The red points represent the sensory data from the configuration space, while the blue points indicate the data obtained from IMU integration. The dashed line denotes the reference teajectory and the sold lines show the ground truth obtained from the depth camera data. From the figure we find that the measurement noise exhibits characteristics of a Gaussian distribution. The variance vector $\bm{\sigma}^2$ of the two sensory body state data are computed using Eq.\ref{eq3}. $\bm{X}_{FPi}$ and $\hat{\bm{X}}_{FPi}$ represent the ground truth and the measured state from sensory data of FP trajectory, respectively. Since the two sets of sensory state data are computed in the same manner, we use $\bm{\sigma}_{imu0}^2$ and $\bm{\sigma}_{JC0}^2$ to distinguish between them. We calculate the variance of measurement errors in position and velocity along the Y and Z directions at time intervals of 0.01 seconds, as shown in Fig\ref{fig_a}.d and Fig\ref{fig_a}.e. It can be observed from the figure that the measurement noise still remains nearly constant over time. Similar results can also be observed in the X direction. Therefore, the covariance matrices and of the two sensory state data of TT II robot can be approximated as constant diagonal matrices $\bm{\sigma}_{imu0}^2$ and $\bm{\sigma}_{JC0}^2$, respectively, and their values can be determined accordingly.
\begin{equation}
	\label{eq3}
	\sigma^2=\frac{1}{n-1}\sum_{i=1}^{n}\left(\bm{X}_{FPi}-\hat{\bm{X}}_{FPi}\right)^2
\end{equation}

\subsubsection{Analysis of Process Noise}
\label{sec2.2.2}
Process noise is typically caused by mechanical clearances, motor backlash, and model inaccuracies. For bipedal robots with the ESVC foot, the elliptical arc length lacks a closed-form solution. To address this, our prior work introduced an approximation calculation based on elementary-function. However, the approximation errors in velocity and position are magnified by the leg length, thereby compromising the accuracy of body state estimation.  Therefore, before designing the state estimator, it is essential to analyze the noise introduced by the ESVC foot and determine its distribution.

\begin{figure}[!t]
	\centerline{\includegraphics[width=\columnwidth]{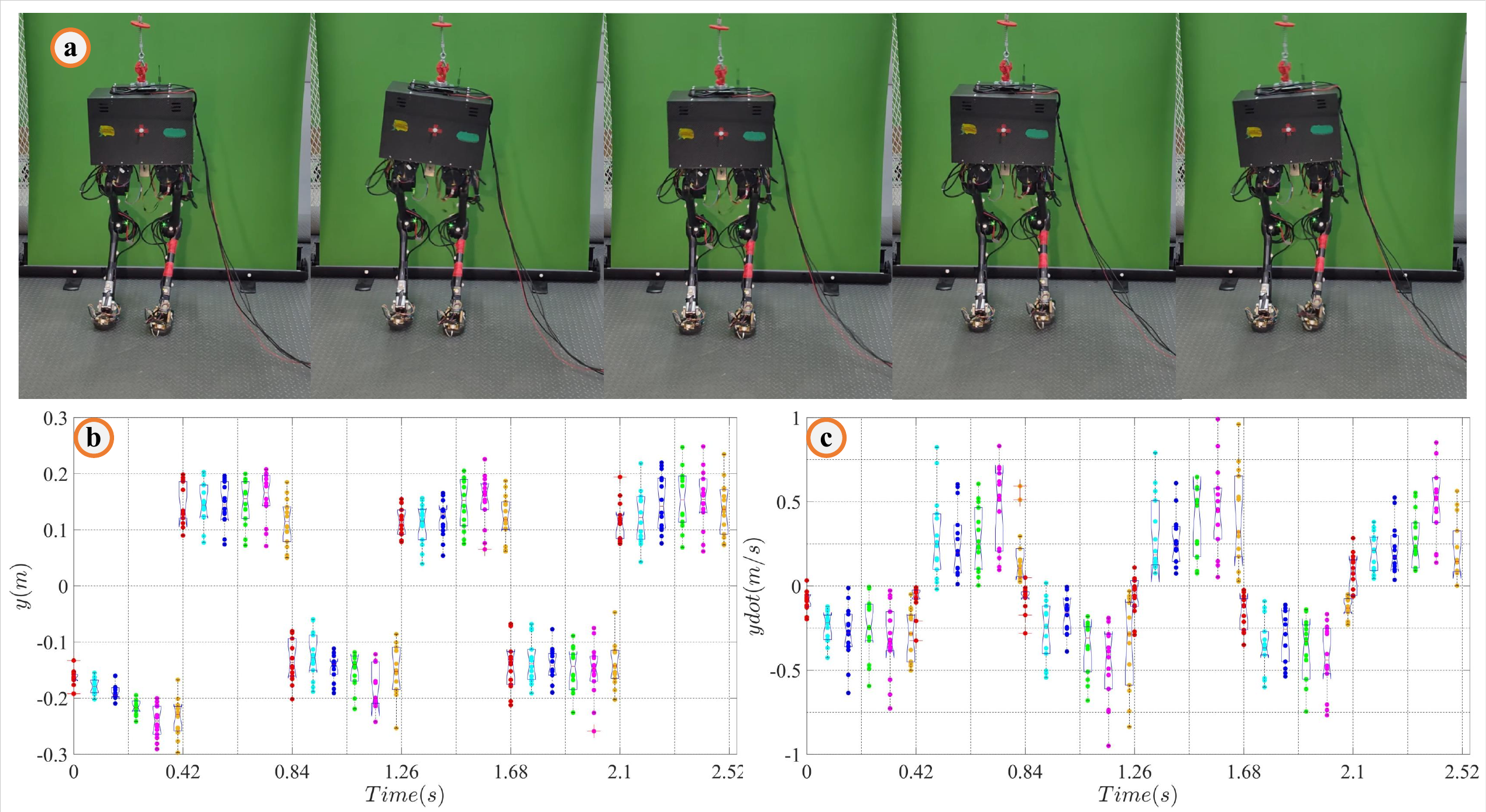}}
	\caption{(a) Snapshots of TT II marking time. (b) FP position trajectories over 16 experiments. (c) (c) FP velocities trajectories over 16 experiments.}
	\label{fig_b}
\end{figure}

Open-loop marking time experiments on the TT II robot were conducted to investigate how modeling errors of the ESVC foot affect the system process noise. We first generate the desired CoM and swing foot target state trajectories using the HLIP model\cite{b25}. Since the ESVC foot operates in coronal plane, the approximate kinematic model is used to formulate contact point position constraints and friction cone constraints. These are incorporated into a full-order trajectory optimization based on DIRCON\cite{b27} to generate the desired joint trajectories. The orientation of swing foot is constrained to maintain parallel with the ground at all times. For HLIP model, the duration of single support phase is 0.42s and the orbit composition is SP1-CP2. The nominal step length during the left support phase(LSP) is –0.303m, which aligns with the relative positions of the left and right feet when the robot is in static standing posture. The target joint positions, velocities, and feedforward torques are obtained by solving the optimization and the robot is controlled to track the trajectories using PD feedforward control and gravity compensation. To ensure experimental repeatability, the TT II robot is initialized to a fixed posture by a calibration device, and then performs marking time by tracking the target joint trajectories, as depicted in Fig\ref{fig_b}.a. Of course, the robot inevitably tilts and falls over time. For analysis, we focus on several steps exhibiting relatively accurate tracking. The reference joint position and data are obtained by the depth camera while the orientation and angular velocity of the support foot are directly measured by an inclinometer and gyroscope mounted at the center of the foot upper plane. The event of support phase transition is determined by the onboard timer. Then the CoM state w.r.t contact point frame is calculated through forward kinematics with URDF file and ESVC foot model(Eq.\ref{eq2}). Sixteen identical experiments were conducted to collect sample data. For clarity, a sampling interval of 0.08 seconds was used. The Y-direction position and velocity of five steps with better performance are plotted in Fig\ref{fig_b}.b and Fig\ref{fig_b}.c, respectively, where the color variation of each point indicates the corresponding sampling time.

It can be observed that even with the ESVC foot, the Y-direction position and velocity of the CoM approximately follow a normal distribution with clearly time-dependent characteristics. Similar results can also be observed in the X and Z directions. Moreover, significant discrepancies exist between the experimental and theoretical trajectories in both velocity and direction, indicating that the TT II robot experiences relatively large process noise. Another noteworthy observation is that, for the Y-direction velocity, the measured data and the theoretical trajectory exhibit opposite signs at each step transition. For instance, the red points at the beginning of the third step in Fig\ref{fig_b}.c correspond to the transition from the RSP and LSP. Theoretically, the FP velocity in the Y direction should be positive at these transitions; however, the feedback data from the robot show a negative velocity, indicating that the swing foot made premature contact with the ground. Nevertheless, since the magnitude of the velocity is small, this indicates a premature contact of the swing foot.

\subsubsection{Regression of Process Noise}
\label{sec2.2.3}
To suppress the influence of incidental fluctuations, we first compute the average trajectory over the six selected steps. As the data storage frequency of robot system is 1kHz, each step for a single experiment will yield 420 data points. We divide the data points into 9 segments: the first eight segments each last 0.05s, and the final segment has a duration of 0.02s. Then the data from the six steps are subsequently merged into a single step, and the error variance trajectory of both CoM state is calculated accordingly. To analyze the error variance, the error variance trajectory is further divided into multiple sub-segments with each containing 20 knots and different weights are assigned to the variance points. Each sub-segment is referred to as a weighted variance window. The first eight segments are each associated with five weighted windows, whereas the final segment contains two.The variance at the center of each window is taken as the representative value of that window. By performing cubic polynomial regression to the representative values of these windows, the characteristics and variations of the process noise error variance within a single step can be clearly observed. 

For a single weighted window, each variance point is assigned a weight based on its time offset from the segment endpoint, and the central value of the window is updated as follows:
\begin{equation}
	\label{eq4}
	\left\{
	\begin{aligned}
		&\bar{\bar{y}}_k=\frac{1}{N}\sum_{n=1}^Nw_{k\_n}y_{k\_n} \\
		&w_{k\_n}= e^-\frac{(t_n-t_k)^2}{2\tau_w^2}\\
	\end{aligned}	
	\right.
\end{equation}
$k$ is the index of current window. $\bar{\bar{y}}_k$ and $y_{k\_n}$ represent the representative value and the other data samples within the window, respectively. $t_k$ and $t_n$ are the corresponding time of $\bar{\bar{y}}_k$ and $y_{k\_n}$. $w_{k\_n}$ is the weights of each point derived from Gaussian kernel function and $\tau_w$ is smoothing parameter.

After processing the data, we used a regression model to extract temporal variation features. For the 9 data segments, the cubic polynomial regression is performed on window center value vector $\bar{\bar{\bm{y}}}_j$. The resulted regression expression can be characterized as:
\begin{equation}
	\label{eq5}
	g_j = a_{j\_1}t^3 + a_{j\_2}t^2 + a_{j\_3}t + a_{j\_3}
\end{equation}
$j$ is the index of the segment and $g_j$ is the fitted value. $t$ is the current sample time. $\bm{a}_j=[a_{j\_1},a_{j\_2},a_{j\_3},a_{j\_4}]^T$ represents the cubic parameters of current segment and can be calculated as:
\begin{equation}
	\label{eq6}
	\bm{a}_j = (\bm{t}_j^T\bm{t}_j)^{-1}\bm{t}_j^T\bar{\bar{\bm{y}}}_j
\end{equation}
$\bm{t}_j$ is the vector containing the sampling times of all window centers within this segment.

\begin{figure}[!t]
	\centerline{\includegraphics[width=\columnwidth]{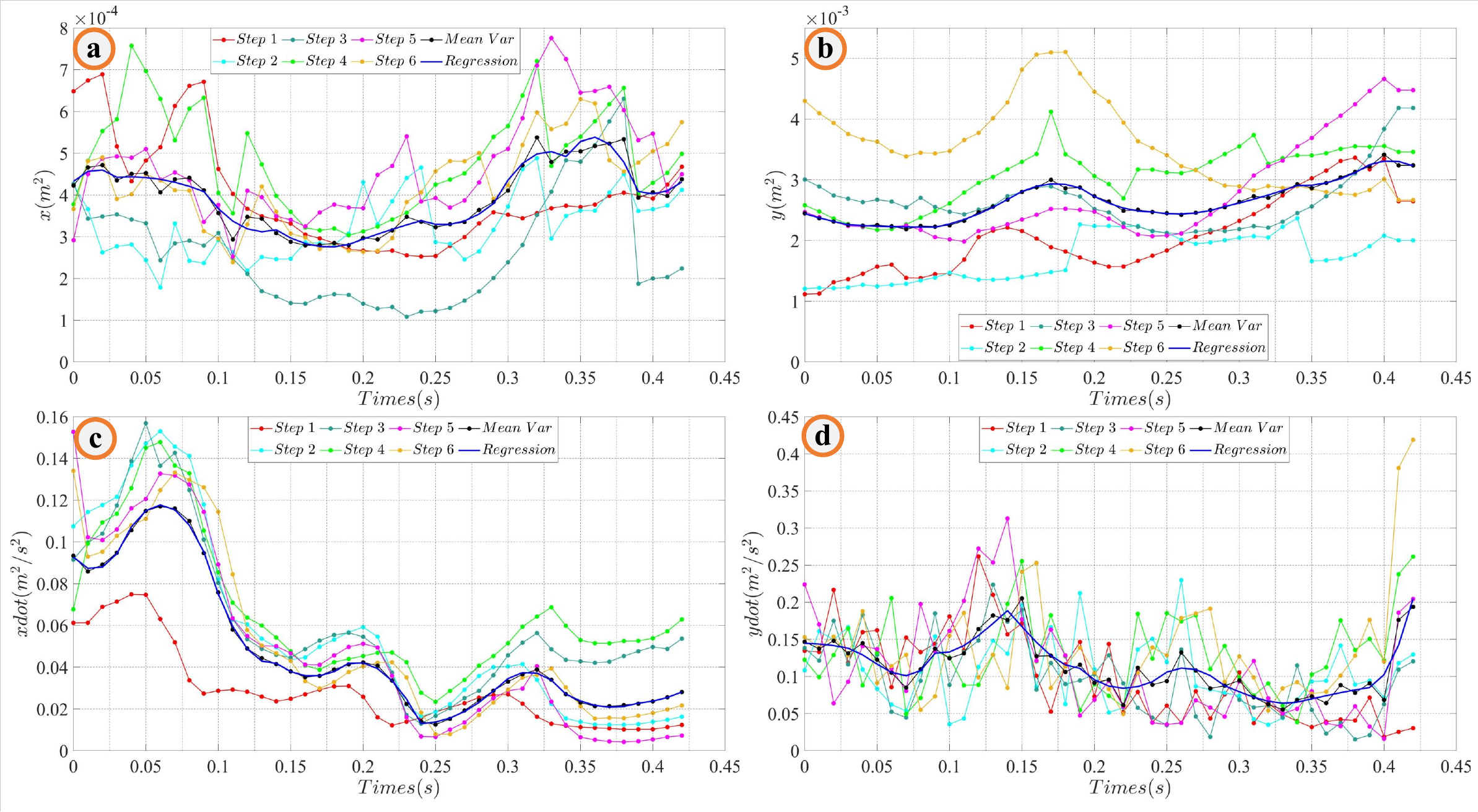}}
	\caption{Process error variance and regression results of TT II walking with ESVC foot over 16 experiments.}
	\label{fig_c}
\end{figure}

For different segments, a first-order continuity constraint is also enforced. Then the process error variances of the CoM position and velocity in both the X and Y directions are depicted in Fig\ref{fig_c}. The black curves denote the average variance of the selected steps and the blue curves represents the regression result. As shown in the figure, the process noise error variance in both the X and Y directions is strongly time-dependent, with trends that are smooth and easily fitted. This observation supports the development of state estimators and performance enhancement methods for robots employing the ESVC foot.

\section{Hierarchical Adaptive Body State Estimator}
\label{sec3}
In the previous section, we analyzed the impact of the ESVC foot on system noise of the TT II robot through physical FP tracking and marking time experiments. For the measurement noise, it exhibited approximately constant error variance over time. For the process noise, we confirmed that despite the modeling errors introduced by the ESVC foot, it still approximately follows a normal distribution. This validates the use of a Gaussian model for characterizing the robot process noise, which is crucial for designing state estimators for bipedal robots with the ESVC foot. Based on these findings, this section presents the proposed adaptive state estimator.

The proposed body state estimator contains two layers. In the first layer, a pre-estimate of the robot sensory data is obtained based on fusion methods. For the second layer, an adaptive variation of extended Kalman filter(EKF) is employed to further update the CoM state with the temporal regression model.

\subsection{Data Fusion Based Pre-Estimation}
\label{sec3.1}
As described in sec\ref{sec2}, the TT II robot provides two types of direct sensory state, $\bm{X}_{imu}$ and $\bm{X}_{JC}$. In this section, we use $\check{\bm{X}_i}=[\check{\bm{p}}_i,\check{\bm{v}}_i]$ to represent the data at $t_i$ after pre-estimation. Due to the significant differences between position and velocity errors, they are treated separately. For position, we directly fuse the two types of sensory position from the robot:
\begin{equation}
	\label{eq7}
	\check{\bm{p}}_i=\bm{p}_{JC\_i}+\bm{K}_{pos}(\bm{p}_{imu\_i}-\bm{p}_{JC\_i})
\end{equation}
$\bm{K}_{pos}$ is a position fusion diagonal matrix composed of $K_{posx}$,  $K_{posy}$ and  $K_{posz}$. The optimal fusion parameter can be easily obtained by Eq.\ref{eq9}:
\begin{equation}
	\label{eq9}
	\bm{K}_{pos\_i} = \frac{\bm{\sigma}_{JC\_i}^2}{\bm{\sigma}_{imu\_i}^2 + \bm{\sigma}_{JC\_i}^2}
\end{equation}
The error variance vector after fusing is $\check{\bm{\sigma}}_{prepos}=[\check{\bm{\sigma}}_{prex},\check{\bm{\sigma}}_{prey},\check{\bm{\sigma}}_{prez}]^T$. Since the two sensory data of the robot are decoupled, the parameters of position fusion are:

\begin{equation}
	\label{eq10}
	\check{\bm{\sigma}}_{prepos}^2=
	\begin{bmatrix}
		(1-K_{posx})^2 & K_{posx}^2\\
		(1-K_{posy})^2 & K_{posy}^2\\
		(1-K_{posz})^2 & K_{posz}^2\\
	\end{bmatrix}
	\begin{bmatrix}
		\bm{\sigma}_{JCpos}^2 & \bm{\sigma}_{imupos}^2
	\end{bmatrix}^T 
\end{equation}

For velocity, since experimental observations indicate that joint velocity encoder errors increase significantly under vibrations,  a two-step data fusion strategy is adopted. First, we construct a differential velocity $\bm{v}_{diff\_i}$ through $\check{\bm{p}}$:
\begin{equation}
	\label{eq11}
	\bm{v}_{diff\_i} = \frac{\check{\bm{p}}_i-\check{\bm{p}}_{i-1}}{\sigma t}
\end{equation}
$\sigma t$ is the sampling interval and the variance of $\bm{v}_{diff\_i}$ can be compute as:
\begin{equation}
	\label{eq12}
	\check{\bm{\sigma}}_{diff}^2=
	\begin{bmatrix}
		(1-K_{posx})^2 & K_{posx}^2\\
		(1-K_{posy})^2 & K_{posy}^2\\
		(1-K_{posz})^2 & K_{posz}^2\\
	\end{bmatrix}
	\begin{bmatrix}
		{\bm{\sigma}_{posJC}^2}^T & {\bm{\sigma}_{posimu}^2}^T 
	\end{bmatrix}
\end{equation}

At first step, we fuse $\bm{v}_{diff\_i}$ and $\bm{v}_{imu\_i}$ as shown in Eq.\ref{eq13}. The corresponding fusion diagonal matrix $\bar{\bm{K}}_{vel}$ and error variance $\bar{\bm{\sigma}}_{vel}^2$ is calculated analogously to $\bm{K}_{pos}$ and $\check{\bm{\sigma}}_{pos}^2$, as illstrated in Eq.\ref{eq13} and Eq.\ref{eq14}.
\begin{equation}
	\label{eq13}
	\left\{
	\begin{aligned}
		&\bar{\bm{v}}_i=\bm{v}_{diff\_i}+\bar{\bm{K}}_{vel}(\bm{v}_{imu\_i}-\bm{v}_{diff\_i})\\
		&\bar{\bm{K}}_{vel}=\frac{\bm{\sigma}_{diff}^2}{\bm{\sigma}_{diff}^2+\bm{\sigma}_{velimu}^2}
	\end{aligned}
	\right.
\end{equation}
\begin{equation}
	\label{eq14}
	\bar{\bm{\sigma}}_{vel}^2=
	\begin{bmatrix}
		(1-K_{velx})^2 & K_{velx}^2\\
		(1-K_{vely})^2 & K_{vely}^2\\
		(1-K_{velz})^2 & K_{velz}^2\\
	\end{bmatrix}
	\begin{bmatrix}
		\bm{\sigma}_{diff}^2 & \bm{\sigma}_{velimu}^2
	\end{bmatrix}^T 
\end{equation}

At second step, the $\bar{\bm{v}}_{i}$ is fused with joint sensory data $\bm{v}_{JC\_i}$ as described in Eq.\ref{eq15}. $\check{\bm{v}}_{jci}$ denotes the resulting velocity and $\check{K}_{vel}$ represents the fusion matrix. The error variance after second fusion $\check{\bm{\sigma}}_{vel}^2$ is depend on $\bar{\bm{\sigma}}_{vel}^2$ and $\bm{\sigma}_{velJC}^2$ , as shown in Eq.\ref{eq16}.
\begin{equation}
	\label{eq15}
	\left\{
	\begin{aligned}
		&\check{\bm{v}}_i=\bar{\bm{v}}_{JC\_i}+\bar{\bm{K}}_{vel}(\bar{\bm{v}}_{i}-\bm{v}_{JC\_i})\\
		&\check{\bm{K}}_{vel}=\frac{\bm{\sigma}_{JC\_i}^2}{\bm{\sigma}_{JC\_i}^2+\bar{\bm{\sigma}}_{vel}^2}
	\end{aligned}
	\right.
\end{equation}
\begin{equation}
	\label{eq16}
	\check{\bm{\sigma}}_{vel}^2=
	\begin{bmatrix}
		(1-K_{velx})^2 & K_{velx}^2\\
		(1-K_{vely})^2 & K_{vely}^2\\
		(1-K_{velz})^2 & K_{velz}^2\\
	\end{bmatrix}
	\begin{bmatrix}
		\bar{\bm{\sigma}}_{vel}^{2} & {\bm{\sigma}_{velJC}}^{2}
	\end{bmatrix}^T
\end{equation}

\subsection{Adaptive post-Estimation}
\label{sec3.2}
For the post-Estimation, the CoM state in X direction is denoted as $\bm{X}_{comx\_i}=[p_{x\_i},v_{x\_i}]^T$ and its state transition equation and observation equation can be written as:
\begin{equation}
	\label{eq17}
	\left\{
	\begin{aligned}
		&\bm{X}_{comx\_i+1}=
		\begin{bmatrix}
			1 & \sigma t & \sigma t^2\\
			0 & 1 & \sigma t
		\end{bmatrix}
		\begin{bmatrix}
			\bm{X}_{comx\_i} & \check{\bm{a}}_{x\_i}+\bm{\omega}_i
		\end{bmatrix}^T
		\\
		&\bm{Z}_{comx\_i+1}=\bm{H}_x
		\begin{bmatrix}
			\check{p}_{comx\_i} & \check{v}_{comx\_i}+\bm{\phi}_i
		\end{bmatrix}^T
	\end{aligned}
	\right.
\end{equation}

$\check{a}_{x\_i}$ is the X direction component of the CoM acceleration $\check{\bm{a}}_{i}$. $\bm{H}_x$ is the observation matrix and $\bm{Z}_{comx\_i+1}$ is the measurement state. $\bm{\omega}_i$ and $\bm{\phi}_i$ represent the process noise and measurement noise, respectively. Before we perform EKF to Eq.\ref{eq17}, the acceleration of CoM, $\check{a}_{x\_i}$ should be determined. Although it is feasible to compute the CoM acceleration directly from foot contact forces and joint accelerations, modeling inaccuracies introduce errors in the results which will be infeasible for integration. Moreover, for the robot without contact force sensor, such as TT II and Cassie, the contact force need additional estimation that also increase the computational burden on the onboard system. It is worth noting that, since the robot is constrained to follow the HLIP model during walking, its CoM acceleration should closely match that of the HLIP behaviour. From another point, as an IMU is located at FP is close to the CoM, the sensor data can also be used to approximate the CoM acceleration. Then the acceleration of CoM can be obtained by fusing the acceleration of HLIP model and FP point:
\begin{equation}
	\label{eq18}
	\check{\bm{a}}_{i}=\bm{a}_{fp\_i}+\bm{K}_a(\bm{a}_{hlip}-\bm{a}_{fp\_i})
\end{equation}

$\bm{a}_{hlip}=\left[ \frac{\tilde{p}_{xi}}{h_0},\frac{\tilde{p}_{yi}}{h_0},0 \right]^T$ is the theoretical acceleration of HLIP. $\tilde{p}_{xi}$ and $\tilde{p}_{yi}$ are the CoM position w.r.t. contact position. $h_0$ is the target CoM height. $\bm{K}_a$ is the fusion diagonal matrix composed of $K_{ax\_i}$,  $K_{ay\_i}$and $K_{az\_i}$. As the CoM state approaches the theoretical model state, the HLIP acceleration becomes more reliable, i.e., $\bm{K}_a$ tends toward 0. Conversely, when the CoM state deviates further from the model state, we place greater trust in the acceleration measured at the FP point. Therefore, a Gaussian kernel function is chosen to determine $K_{ax\_i}$. Taking the X-direction as an example:
\begin{equation}
	\label{eq19}
	K_{ax\_i}=e^{\frac{(\tilde{p}_{x\_i}-p_{hx\_i})^2+(\tilde{v}_{x\_i}-v_{hx\_i})^2}{2\tau_a^2}}
\end{equation}
$p_{hx\_i}$ and $v_{hx\_i}$ is the desired CoM state of HLIP, $\tau_a$ is the smoothing parameter. At this point, both the state transition equation and observation equation in Eq.\ref{eq17} are fully defined.

As discussed in Sec\ref{sec2}, both measurement noise and process noise are approximately follow a Gaussian distribution and the error covariance matrices can be pridicted at any time through the temporal regression model. It is important to note that the regression model is established based on the FP trajectory tracking task and the marking time task. In practice, when the robot changes its behavior, such as variations in walking speed or CoM height, the characteristics of both measurement noise and process noise should also change. Consequently, the error covariance matrices also need to be updated in time. In other words, it is impractical to analyze the noise characteristics for all possible robot motions and the post-estimation is required to adaptively upate the covariance matrices. For bipedal robots, squatting and walking serve as templates for more complex motions. Therefore, the regression model derived from noise analysis can serve as an initial estimate for updating the current covariance matrices, and the subsequent updates will rollover with the historical covariance matrices. The measurement noise covariance and process noise covariance are updated as:
\begin{equation}
	\label{eq20}
	\left\{
	\begin{aligned}
		&\tilde{\bm{R}}_{C\_i}=\bm{\alpha} \bm{R}_{C\_i} + (\bm{I}-\bm{\alpha})\hat{\bm{R}}_{C\_i}\\
		&\tilde{\bm{Q}}_{C\_i}=\bm{\beta}_0\bm{Q}_{C\_i} + \bm{\beta}_1\hat{\bm{Q}}_{C\_i} + \bm{\beta}_2\tilde{\bm{Q}}_{C\_i-1}
	\end{aligned}
	\right.
\end{equation}

$\tilde{\bm{R}}_{C\_i}$ and $\tilde{\bm{Q}}_{C\_i}$ are the updated error covariance matrices of measurement noise and process noise, respectively, while $\bm{R}_{C\_i}$ and $\bm{Q}_{C\_i}$ represent the initial estimate covariance matrices which are initialized by $\bm{R}_{C\_0}$ and $\bm{Q}_{C\_0}$ from Sec\ref{sec2}.  $\hat{\bm{R}}_{C\_i}$ and $\hat{\bm{Q}}_{C\_i}$ are estimated from the variance of residuals. For simplicity, we directly use the squared residuals. For the temporal characteristics of the process noise, the covariance matrix from the previous time step is used as the third term in the update loop. $\alpha\in \mathbb{R}^{6\times6}$ is a constant measurement update parameter matrix, corresponding to the three-dimensional position and velocity of the CoM, $\bm{\beta}_0$, $\bm{\beta}_1$ and $\bm{\beta}_2$ are the process update parameter matrix which denotes the weights of initial estimate covariance matrices, measurement residuals and historical covariance matrix, respectively. And these matrices satisfy the constraint  $\bm{\beta_0}+\bm{\beta_1}+\bm{\beta_2}=\bm{I}_6$. At this point, standard EKF operations can be applied to Eq.\ref{eq17}, which are not elaborated here. Then the error variance $\bm{\sigma}_{JC\_i}^2$ and $\bm{\sigma}_{imu\_i}^2$ of sensory data still need to be updated, as shown in Eq.\ref{eq21}. $\bm{e}_{JC\_i}$ denotes the difference between CoM state $\bm{X}_{JC\_i}$ derived from joint sensory data and the post-estimated state $\tilde{\bm{X}}_{i}$ and the update of $\bm{\sigma}_{imu_i}^2$ follows the same procedure as that of $\bm{\sigma}_{JC_i}^2$. At the beginning of robot motion, $\bm{\sigma}_{imu_i}^2$ and $\bm{\sigma}_{JC_i}^2$ are initialized using $\bm{\sigma}_{imu0}^2$ and $\bm{\sigma}_{JC0}^2$, respectively.
\begin{equation}
	\label{eq21}
	\left\{
	\begin{aligned}
		&\bm{\sigma}_{JC\_i+1}^2=\frac{1}{n-1}\sum_{n=1}^{N}\bm{e}_{JC\_i}-\bar{\bm{e}}_{JC\_i}\\
		&\bm{e}_{JC\_i}=X_{JC\_i} - \tilde{X}_i
	\end{aligned}
	\right.
\end{equation}
\begin{figure}[!t]
	\centerline{\includegraphics[width=\columnwidth]{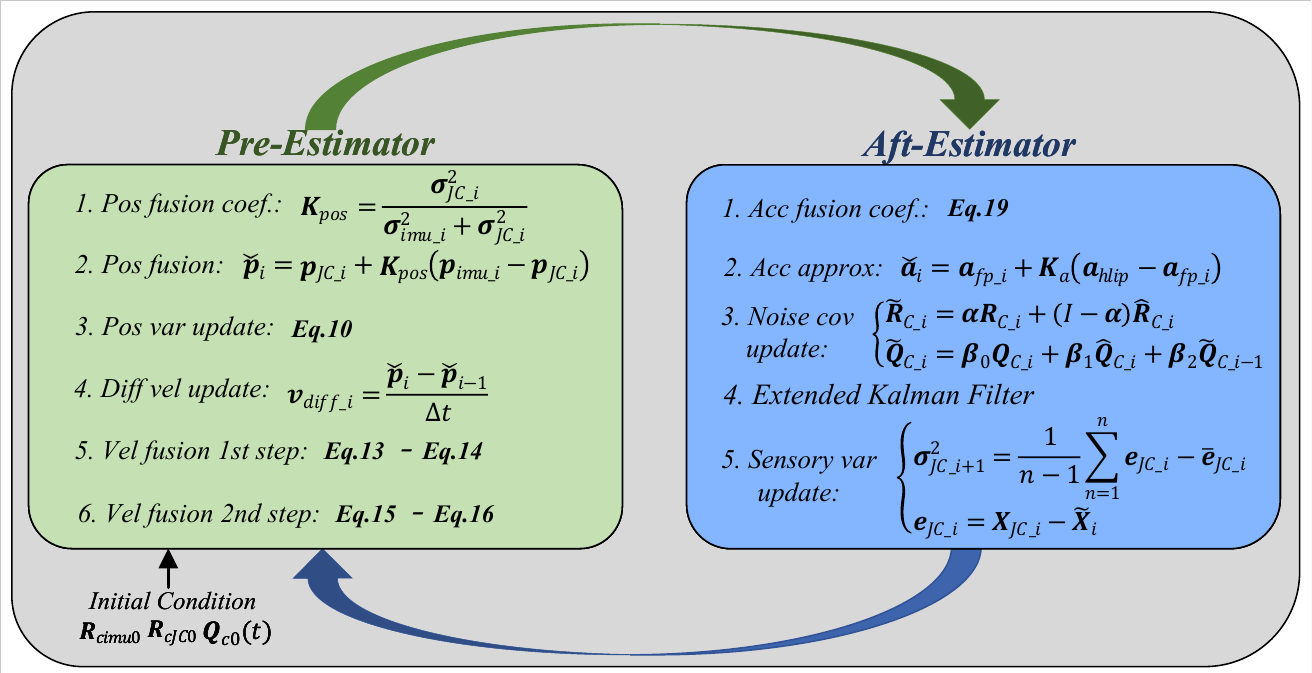}}
	\caption{The overall architecture of the proposed hierarchical body state estimator.}
	\label{fig_d}
\end{figure}

For clarity, the proposed hierarchical body state estimator is illustrated in Fig\ref{fig_d}. Components with a green background represent the data fusion-based pre-estimation stage, while those with a blue background correspond to the adaptive post-estimation.

\section{Experiment Evaluation}
\label{sec4}
\subsection{Experimental Setup}
\label{sec4.1}
To evaluate the proposed body state estimator, two types of physical experiments were conducted on the TT II robot: marking time and variable-speed lateral walking. The joints of TT II are actuated by AK-8064, a high-power motor designed by Cubemars\cite{b28}. The onboard computer is an Intel i7-8665U processor running Xenomai 3.2 real-time Linux as the main controller. The robot communication network is based on CAN network and all the sensory data are returned by HWT-9073-CAN developed by WIT MOTION\cite{b29}. Each ESVC foot is equipped with four pressure sensors for contact detection. The robot is controlled to follow the HLIP model, with the swing foot trajectory generated using cubic Hermite spline. The target joint space trajectory is synthesized through optimization-based phased inverse kinematics(PIK) and solved by the DRAKE robotics toolbox\cite{b30}. The actuated joints track the reference trajectory via PD control with feedforward compensation.

\subsection{Marking Time}
\label{sec4.2}
\begin{figure}[!t]
	\centerline{\includegraphics[width=\columnwidth]{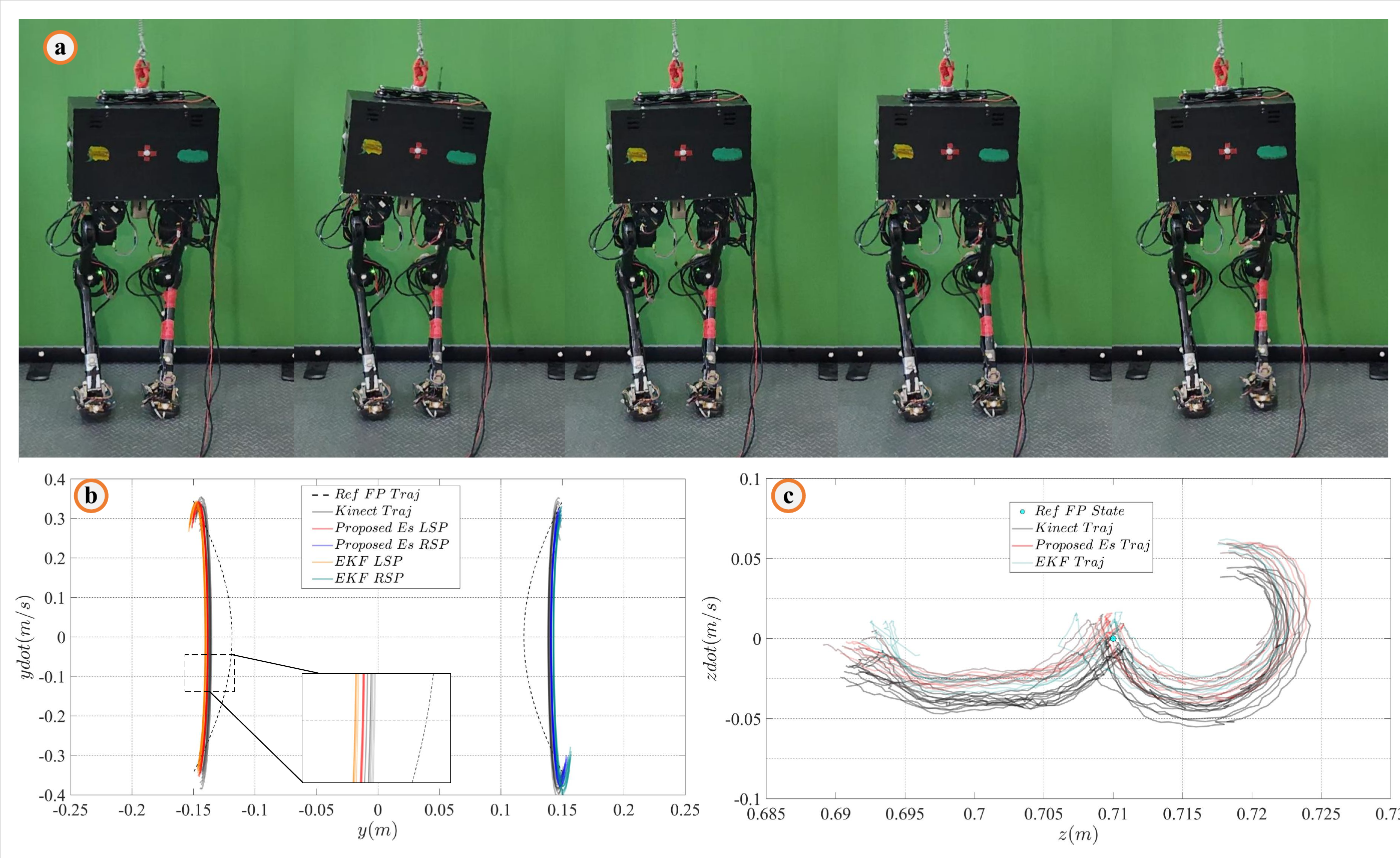}}
	\caption{(a). The snapshots of marking time experiment with TT II;\ (b). CoM phase portrait of Y-direction;\ (c). CoM phase portrait in Z-direction.}
	\label{fig_e}
\end{figure}

To verify the accuracy of the estimator, marking time experiment was conducted on the TT II robot and the snapshots of a complete step is shown in Fig\ref{fig_e}.a. Acquiring the true states of a dynamic system is inherently challenging and an approximate ground truth is established by capturing joint kinematic data with a depth camera and reflective markers. Then the CoM trajectory using forward kinematics derived from the URDF model. The CoM phase portrait of Y-direction and Z-direction are shown in Fig\ref{fig_e}.b and Fig\ref{fig_e}.c, respectively. The black dashed line represents the reference trajectory of the HLIP model, while the dark gray solid line denotes the ground truth of the walking center of mass. The red and blue curves represent the feedback data estimated by the proposed method, corresponding to the left support phase and the right support phase. From the figure, it can be observed that the CoM states estimated by the proposed method closely approximate the ground truth during marking time. However, due to inherent measurement errors, the vision-based feedback also deviates to some extent from the true values. To provide a comparative baseline, we also introduce the EKF, as indicated by the orange and sky blue curves in Fig\ref{fig_e}.b and Fig\ref{fig_e}c. Through comparison, it is observed that the proposed estimator can stably provide accurate state estimates during robot locomotion, with slightly better performance than the EKF-based method.

\subsection{Variable-speed Lateral Walking}
\label{sec4.3}
\begin{figure}[!t]
	\centerline{\includegraphics[width=\columnwidth]{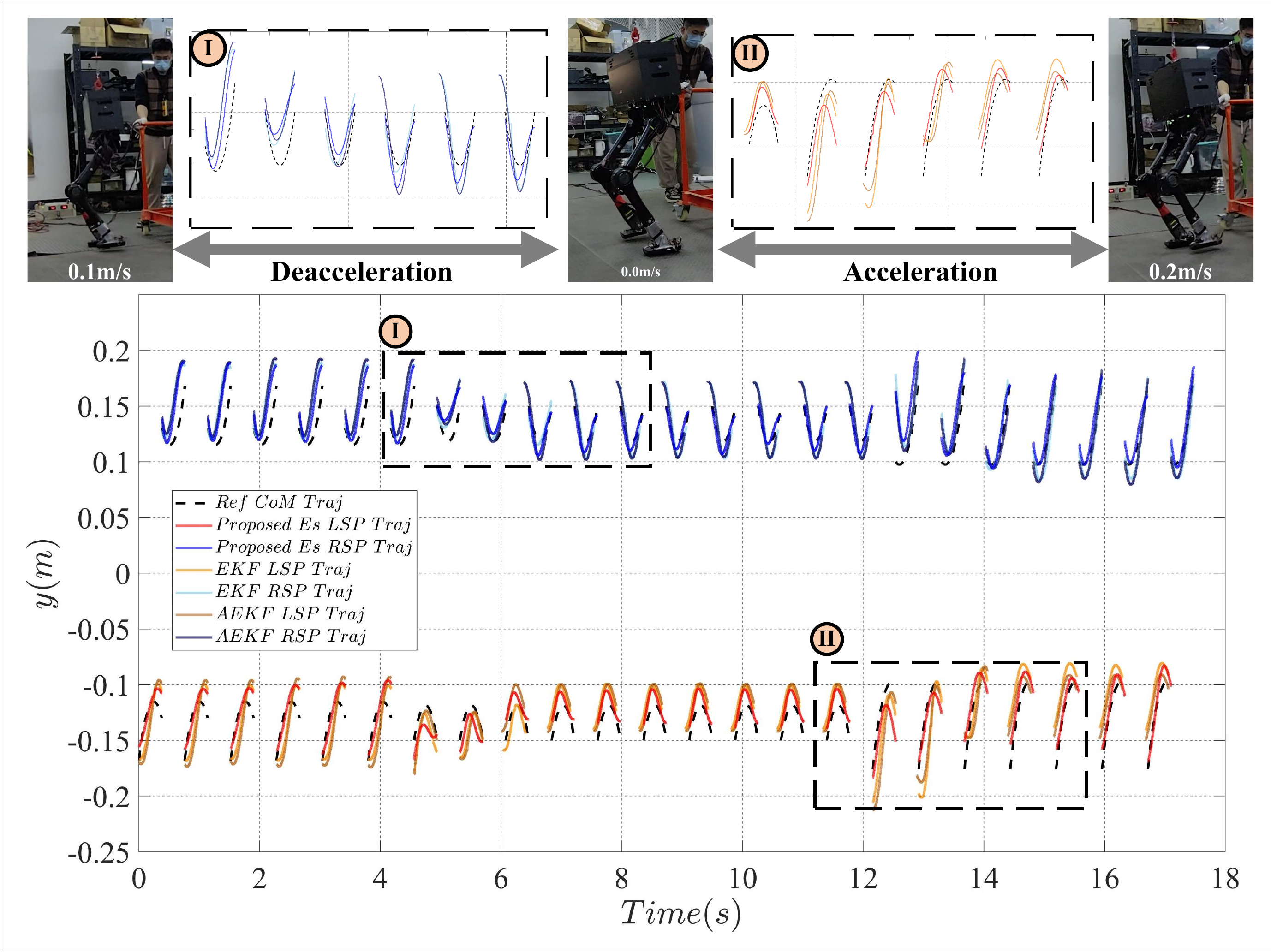}}
	\caption{Performance evaluation of EKF, AEKF and proposed estimator through CoM position in Y-direction.}
	\label{fig_f}
\end{figure}

To evaluate the performance of the state estimator under more complex locomotion conditions, a variable-speed lateral walking experiment was conducted. The TT II robot was controlled to decelerate from lateral walking at 0.1m/s to marking time, and then accelerate to a lateral walking speed of 0.2m/s, as illustrated in Fig\ref{fig_f}. Given the increased difficulty in obtaining accurate ground truth during dynamic walking, the EKF-based estimation is adopted as the reference baseline for comparison in this section. To further evaluate the adaptability of the proposed state estimator, the Adaptive EKF(AEKF) is also included as a baseline for comparison. The number of steps required for the robot to reach stable walking is used as the evaluation criterion. To highlight the influence of the ESVC foot, the variation of the CoM position in the Y-direction over time is plotted in Fig\ref{fig_f}. The red and blue curves represent the state estimates obtained by the proposed method during the LSP and RSP phases, respectively. The EKF results are shown in orange and sky blue, while the Adaptive EKF (AEKF) estimates are depicted in brown and navy blue. Through observation, we find that during steady walking, all three estimators exhibit comparable accuracy, and their outputs are sufficient to enable stable walking of the robot equipped with the ESVC foot. In the acceleration and deceleration phases, the convergence performance of the three estimators varies significantly. The trajectories corresponding to the acceleration and deceleration phases are magnified and presented in the upper section of Fig\ref{fig_f}. In the deceleration phase, the HLIP theoretical model plans the transition within two steps. In contrast, the robot employing the proposed estimator completed the deceleration over four steps, primarily due to modeling inaccuracies and residual estimation errors relative to the true states. The AEKF-based method also completed the deceleration during the fourth LSP, whereas the EKF required until the sixth RSP step to complete the transition. This indicates that both the proposed state estimator and the AEKF exhibit faster convergence than the EKF. In the acceleration phase, the robot using the proposed estimator reached the target stable state also within four steps, while both the AEKF and EKF required six steps to achieve stability. This demonstrates that the proposed estimator achieves faster convergence and, consequently, exhibits better adaptability compared to both EKF and AEKF.

\section{Conclusion}
\label{sec5}
This paper investigates body state estimation for bipedal robots equipped with an ESVC (Ellipse-based Segmental Varying Curvature) foot—a bioinspired design that improves gait efficiency but introduces modeling uncertainties due to complex foot–ground interactions. Physical experiments are conducted to determine the distribution of measurement and process noise, and a time-varying fitting model of covariance matrix is constructed via weighted cubic polynomial regression. A hierarchical adaptive estimator is proposed, consisting of a multi-sensor data fusion–based pre-estimation stage and an adaptive post-estimation stage. In physical marking-time and variable-speed lateral walking, the proposed state estimation method delivers performance comparable to that of the AEKF and slightly better than the EKF. Moreover, when the robot's motion changes, the proposed estimator demonstrates superior adaptability compared to both EKF and AEKF.

\end{document}